\documentclass[10pt,twocolumn,letterpaper]{article}

\usepackage{cvpr}
\usepackage{times}
\usepackage{epsfig}
\usepackage{graphicx}
\usepackage{amsmath}
\usepackage{amssymb}
\usepackage{mdwlist}

\newcommand*{\MYHEAD}{} 

\usepackage[ruled,titlenotnumbered,linesnumbered,noline]{algorithm2e}

\providecommand{\DontPrintSemicolon}{\dontprintsemicolon}

\newcommand{\exclude}[1]{}

\usepackage[pagebackref=true,breaklinks=true,letterpaper=true,colorlinks,bookmarks=false]{hyperref}

\cvprfinalcopy 

\def\cvprPaperID{1710} 

\ifcvprfinal\fi
\begin{document}

\ifdefined\MYHEAD
    \pagestyle{myheadings}
    \setlength{\headsep}{0.4in}
    \markright{\small L. Gorelick, Y. Boykov, O. Veksler, I. Ben Ayed, A. Delong, arXiv:1311.1856v2, April 2014 
\hspace{1ex} (to appear at CVPR 2014)  \hfill p. }
   \setcounter{page}{1}  
\else
\fi

\title{Submodularization for Quadratic Pseudo-Boolean Optimization}

\author{Lena Gorelick \hspace{2ex} Yuri Boykov \hspace{2ex} Olga Veksler \\
Computer Science Department \\
University of Western Ontario
 \and    \hspace{-1ex}
Ismail  Ben Ayed\\  \hspace{-1ex}
GE Healthcare\\
\and   \hspace{-2ex}
Andrew Delong \\  \hspace{-2ex}
Elect. \& Comp. Engineering \\ \hspace{-2ex}
University of Toronto
}

\maketitle
\ifdefined\MYHEAD
    \thispagestyle{myheadings}
\else
    \thispagestyle{empty}
\fi

\begin{abstract}
Many computer vision problems require optimization of binary non-submodular energies.
We propose a general optimization framework based on {\em local submodular approximations} (LSA).
Unlike standard LP relaxation methods that linearize the whole energy globally,
our approach iteratively approximates the energies locally.
On the other hand, unlike standard local optimization methods (\eg gradient descent or projection techniques)
we use non-linear submodular approximations and optimize them without leaving the domain of integer solutions.
We discuss two specific LSA algorithms based on {\em trust region}  and {\em auxiliary function} principles,
LSA-TR and LSA-AUX. These methods obtain state-of-the-art results on a wide range of applications outperforming 
many standard techniques such as LBP, QPBO, and TRWS.   
While our paper is focused on pairwise energies, our ideas extend to higher-order problems. The code is available online \footnote{\url{http://vision.csd.uwo.ca/code/}}.
\end{abstract}


\section{Introduction} \label{sec:intro}

We address a general class of binary pairwise non-submodular energies, which are widely used in
applications like segmentation, stereo, inpainting, deconvolution, and many others.
Without loss of generality, the corresponding binary energies can be transformed into the form\footnote{Note that 
such transformations are up to a constant.}
\begin{equation} \label{eq:en}
E(S) = S^TU + S^T M S, \;\;\;\;\;\;S\in \{0,1\}^{\Omega}
\end{equation}
where $S=\{s_p\,|\,p\in\Omega\}$ is a vector of binary indicator variables defined on pixels $p\in\Omega$, vector
$U=\{u_p\in{\cal R}\,|\,p\in\Omega\}$ represents unary potentials, and symmetric matrix $M=\{m_{pq}\in{\cal R}\,|\,p,q\in\Omega\}$ 
represents pairwise potentials. Note that in many practical applications matrix $M$ is sparse since elements $m_{pq}=0$
for all non-interacting pairs of pixels. We seek solutions to the following integer quadratic optimization problem 
\begin{equation}\label{eq:iqp}
\min_{S\in \{0,1\}^{\Omega}} E(S).
\end{equation}
When energy  \eqref{eq:en} is {\em submodular}, \ie $m_{pq}\leq 0\;\;\forall (p,q)$,
globally optimal solution for  \eqref{eq:iqp} can be found in a low-order polynomial 
time using graph cuts \cite{Boros01pseudo-booleanoptimization}. The general non-submodular case of problem  \eqref{eq:iqp} is NP hard.

\subsection{Standard linearization methods} 

Integer quadratic programming problems is a well-known challenging optimization problem with extensive literature in the combinatorial 
optimization community, \eg see \cite{lazimy:82,goemans:95,Boros01pseudo-booleanoptimization}. 
It often appears in computer vision where it can be addressed with many methods including spectral and 
semi-definite programming relaxations, \eg see \cite{olsson:CVIU08,Keuchel-et-al-02a}. 

\begin{figure}
\centering
\begin{tabular}{c@{\hskip 5ex}c}
\includegraphics[width =0.4\columnwidth]{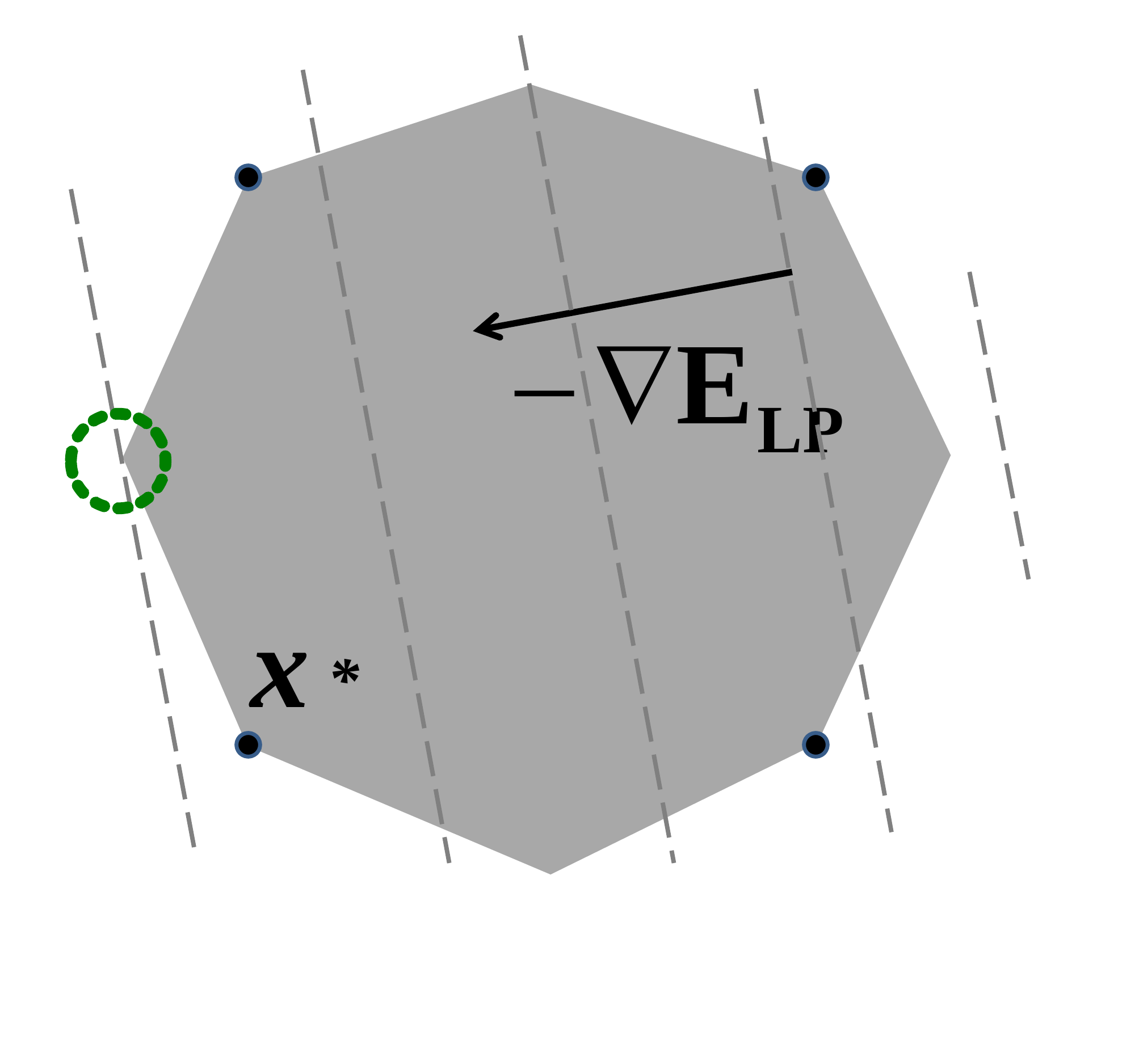} &
\includegraphics[width =0.4\columnwidth]{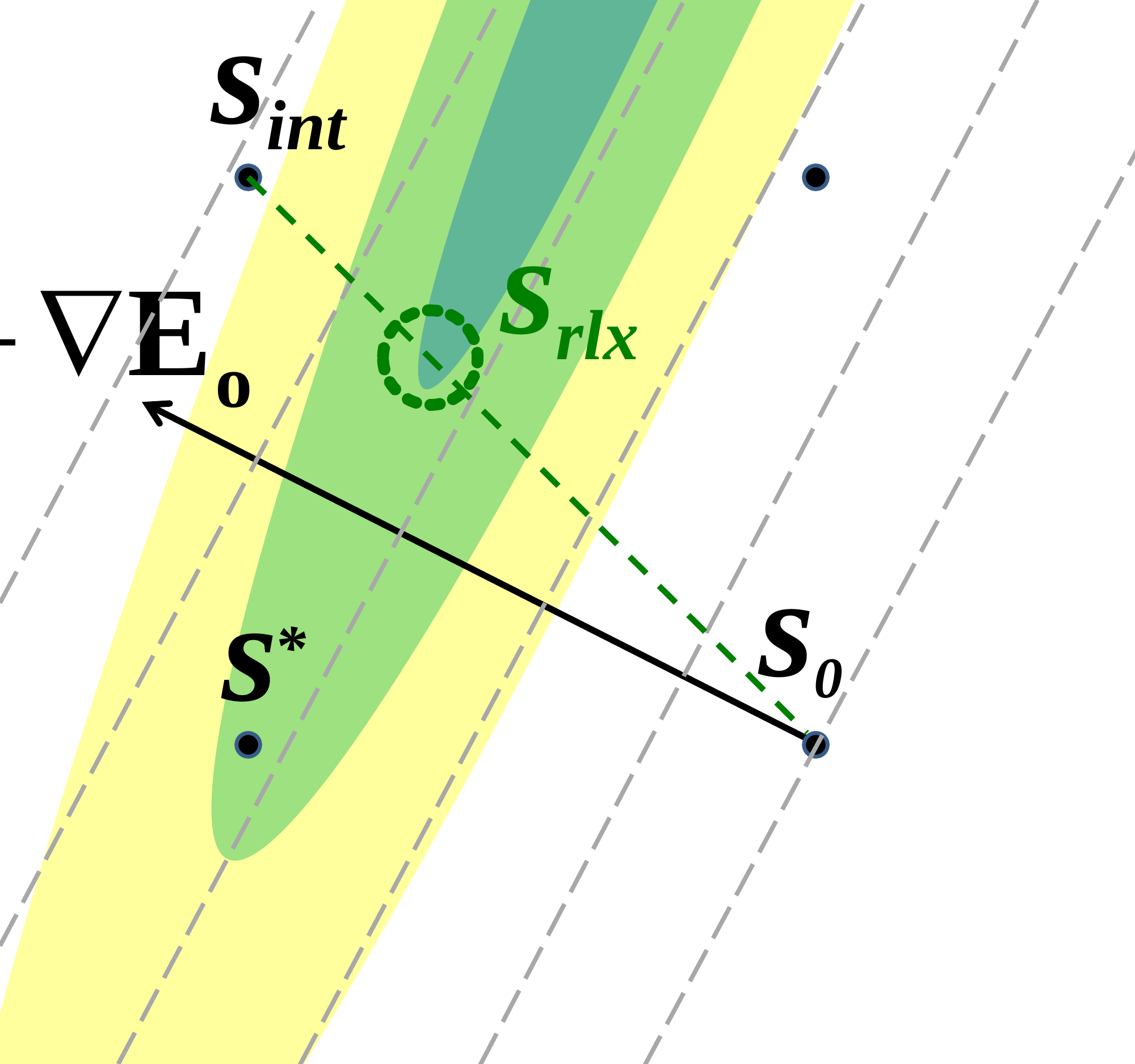} \\
(a) global linearization & (b) local linearization
\end{tabular}
\caption{Standard linearization approaches for \eqref{eq:en}-\eqref{eq:iqp}. Black dots are
integer points and $*$ corresponds to the global optimum of \eqref{eq:iqp}. 
Colors in (b) show iso-levels of the quadratic energy \eqref{eq:en}. 
This energy can be linearized by introducing additional variables and linear constraints, see
a schematic polytope in (a) and \cite{wainwright:03}. 
Vector $\nabla E$ is the gradient of the global linearization of \eqref{eq:en}  in (a) and 
the gradient of the local linear approximation of \eqref{eq:en} at point $S_0$ in (b).    
\label{fig:overview}} 
\end{figure}

Methods for solving \eqref{eq:iqp} based on LP relaxations, 
\eg QPBO \cite{rother-et-al-cvpr-2007} and TRWS \cite{GTRWS:arXiv12},
are considered among the most powerful in computer vision \cite{kappes-2013}. 
They approach integer quadratic problem \eqref{eq:iqp} by {\em global linearization} of the objective function 
at a cost of introducing a large number of additional variables and linear constraints.
These methods attempt to optimize the relaxed LP or its dual. However, 
the integer solution can differ from the relaxed solution circled in Fig.\ref{fig:overview}(a).
This is a well-known {\em integrality gap} problem. 
Most heuristics for extracting an integer solution from the relaxed solution 
have no {\em a priori} quality guarantees.

Our work is more closely related to {\em local linearization} techniques for approximating \eqref{eq:iqp}, \eg
parallel ICM, IPFP \cite{NIPS09LeordeanuHS09}, and other similar methods \cite{NIPS10BrendelTodorovic}.
Parallel ICM iteratively linearizes energy $E(S)$ around current solution $S_0$ using Taylor expansion and  makes 
a step by computing an integer minimizer $S_{int}$ of the corresponding linear approximation, see Fig.\ref{fig:overview}(b). 
However, similarly to Newton's methods, this approach often gets stuck in bad local minima by making 
too large steps regardless of the quality of the approximation. 
IPFP attempts to escape such minima by reducing the step size. It explores the continuous line between integer minimizer 
$S_{int}$ and current solution $S_0$ and finds optimal relaxed solution $S_{rlx}$ with respect to the original quadratic energy.
Similarly to the global linearization methods, see Fig.\ref{fig:overview}(a), such continuous solutions
give no quality guarantees with respect to the original integer problem \eqref{eq:iqp}.

\subsection{Overview of submodularization}

Linearization has been a popular approximation approach to integer quadratic problem \eqref{eq:en}-\eqref{eq:iqp}, 
but it often requires relaxation leading to the integrality gap problem. 
We propose a different approximation approach, which we refer to as {\em submodularization}. 
The main idea is to use submodular approximations of energy \eqref{eq:en}. 
We propose several approximation schemes that keep submodular terms in \eqref{eq:en} 
and linearize non-submodular potentials in different ways leading to very different 
optimization algorithms. Standard {\em truncation} of non-submodular pairwise terms\footnote{ 
Truncation is known to give low quality results, e.g. Fig.\ref{fig:repulsion}, Tab.\ref{tab:chinese}.} 
and some existing techniques for high-order energies  \cite{FTR:cvpr13,Bilmes2005,rother:cvpr06,AuxCut:cvpr13} 
can be seen as {\em submodularization} examples, as discussed later.
Common properties of submodularization methods is that they compute globally optimal 
integer solutions of the approximation and do not need to leave the domain of discrete solutions 
avoiding integrality gaps. Sumbodularization can be seen as a generalization of local linearization methods
since it uses more accurate higher-order approximations.

One way to linearize non-submodular terms in \eqref{eq:en} is to compute their Taylor expansion 
around current solution $S_0$. Taylor's approach is similar to IPFP  \cite{NIPS09LeordeanuHS09}, 
but they linearize all terms including submodular ones. In contrast to IPFP, our overall approximation of $E(S)$ at $S_0$ 
is not linear; it belongs to a more general class of submodular functions. Such non-linear approximations are more accurate 
while still permitting efficient optimization in the integer domain.  

We also propose a different mechanism for controlling the step size. Instead of exploring relaxed solutions on 
continuous interval $[S_0,S_{int}]$ in Fig.\ref{fig:overview}b, we compute integer intermediate solutions $S$ by
minimizing local submodular approximation over $\{0,1\}^\Omega$ under additional distance constraints $||S-S_0||<d$. 
Thus, our approach avoids integrality gap issues. 
For example, even linear approximation model in Fig.\ref{fig:overview}b can produce solution $S^*$ 
if Humming distance constraint $||S-S_0||\leq 1$ is imposed.
This local submodularization approach to \eqref{eq:en}-\eqref{eq:iqp} fits a general {\em trust region framework} 
\cite {fletcher:87,TRreview:Yuan,olsson:CVIU08,FTR:cvpr13} and we refer to it as LSA-TR.

Our paper also proposes a different local submodularization approach to \eqref{eq:en}-\eqref{eq:iqp} based on the general 
{\em auxiliary function} framework \cite{Lange2000,Bilmes2005,AuxCut:cvpr13}\footnote{{\em Auxiliary functions} 
are also called {\em surrogate functions} or {\em upper-bounds}. The corresponding approximate
optimization technique is also known as the {\em majorize-minimize} principle \cite{Lange2000}. }.  Instead of Taylor expansion, 
non-submodular terms in $E(S)$ are approximated by linear upper bounds specific to current solution $S_0$.
Combining them with submodular terms in $E(S)$ gives a submodular upper-bound approximation, 
a.k.a. an {\em auxiliary function}, for $E(S)$ that can be globally minimized within integer solutions.
This approach does not require to control the step sizes as the global minimizer of an auxiliary function is guaranteed to
decrease the original energy $E(S)$. Throughout the paper we refer to this type of 
local submodular approximation approach as LSA-AUX. 

Some auxiliary functions were previously proposed in the context of high-order energies \cite{Bilmes2005,AuxCut:cvpr13}.
For example,  \cite{Bilmes2005} divided the energy into submodular and supermodular parts and replaced the 
latter with a certain permutation-based linear upper-bound. The corresponding auxiliary function allows 
polynomial-time solvers. However, experiments in \cite{rother:cvpr06} (Sec. 3.2)  demonstrated
limited accuracy of the permutation-based bounds \cite{Bilmes2005} on high-order segmentation problems.
Recently, Jensen inequality was used in \cite{AuxCut:cvpr13} to derive linear upper bounds for several important 
classes of high-order terms that gave practically useful approximation results. Our LSA-AUX method is first 
to apply auxiliary function approach to arbitrary (non-submodular) pairwise energies. 
We discuss all possible linear upper bounds for pairwise terms and study several specific cases. 
One of them corresponds to the permutation bounds \cite{Bilmes2005} and is denoted by LSA-AUX-P.

Recently both {\em trust region} \cite {fletcher:87,TRreview:Yuan,olsson:CVIU08} and  
{\em auxiliary function} \cite{Lange2000} frameworks  proved 
to work well for optimization of energies with high-order regional terms \cite{FTR:cvpr13,AuxCut:cvpr13}. 
They derive specific linear \cite{FTR:cvpr13} or upper bound  \cite{AuxCut:cvpr13} approximations for 
non-linear cardinality potentials, KL and other distances between segment and target appearance models.
To the best of our knowledge, we are the first to develop trust region and auxiliary function methods for
integer quadratic optimization problems \eqref{eq:en}-\eqref{eq:iqp}. 


{\bf Our contributions} can be summarized as follows:
\vspace{-1ex}
\begin{itemize*}
\item A general {\em submodularization} framework for solving integer quadratic optimization problems \eqref{eq:en}-\eqref{eq:iqp}
based on {\em local submodular approximations} (LSA). Unlike global linearization methods, LSA constructs an approximation model 
without additional variables. Unlike local linearization methods, LSA uses a more accurate approximation functional.
\item In contrast to the majority of standard approximation methods, LSA avoids integrality gap issue by working strictly
within the domain of  discrete solutions. 
\item State-of-the-art results on a wide range of applications. Our LSA algorithms outperform
QPBO, LBP, IPFP, TRWS, its latest variant SRMP, and other standard techniques for \eqref{eq:en}-\eqref{eq:iqp}.
\end{itemize*}

\section{Description of LSA Algorithms}\label{sec:overview}

In this section we discuss our framework in detail. Section \ref{sec:tr} derives local submodular approximations and describes how to incorporate them in the trust region framework. Section \ref{sec:aux} briefly reviews auxiliary function framework and shows how to derive local auxiliary bounds.

\begin{figure*}
\centering
\begin{tabular}{c@{\hskip 1ex}c@{\hskip 1ex}c}
\includegraphics[width =0.32\textwidth]{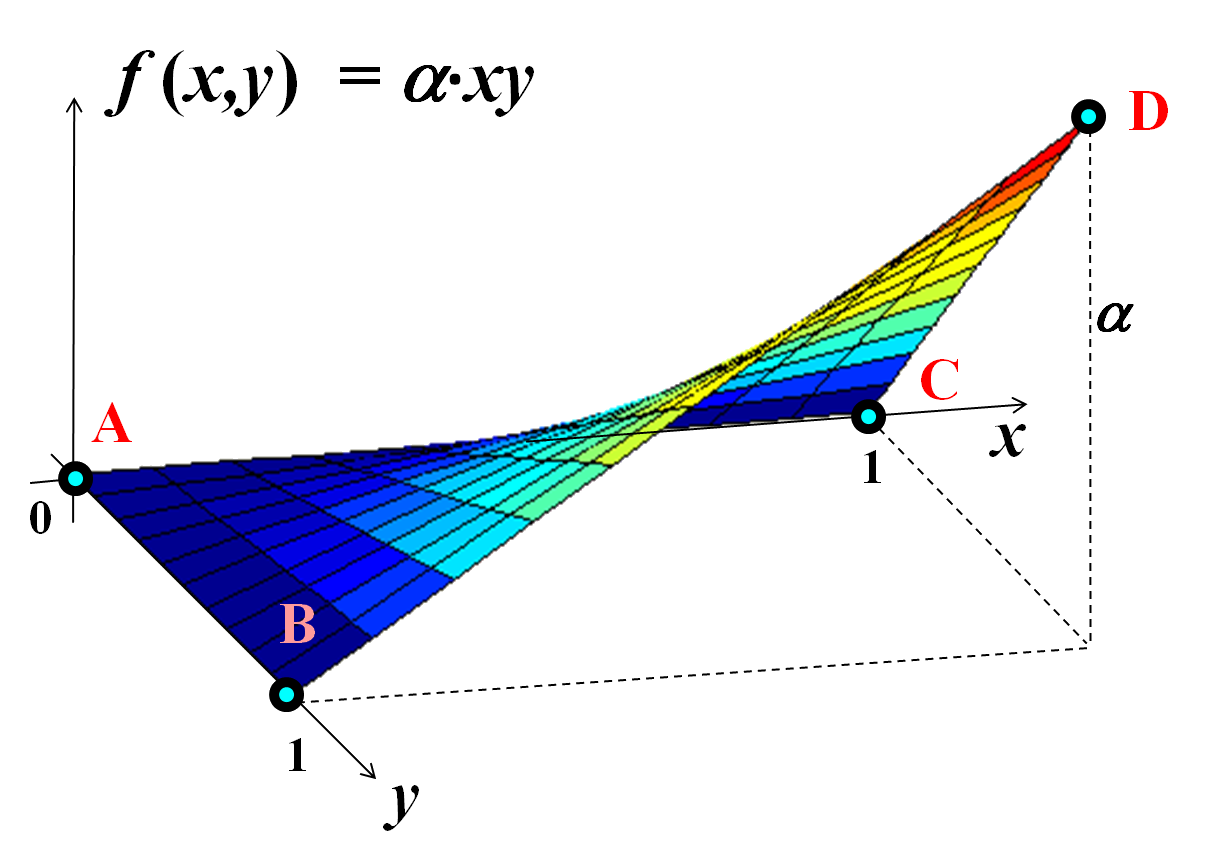} &
\includegraphics[width =0.32\textwidth]{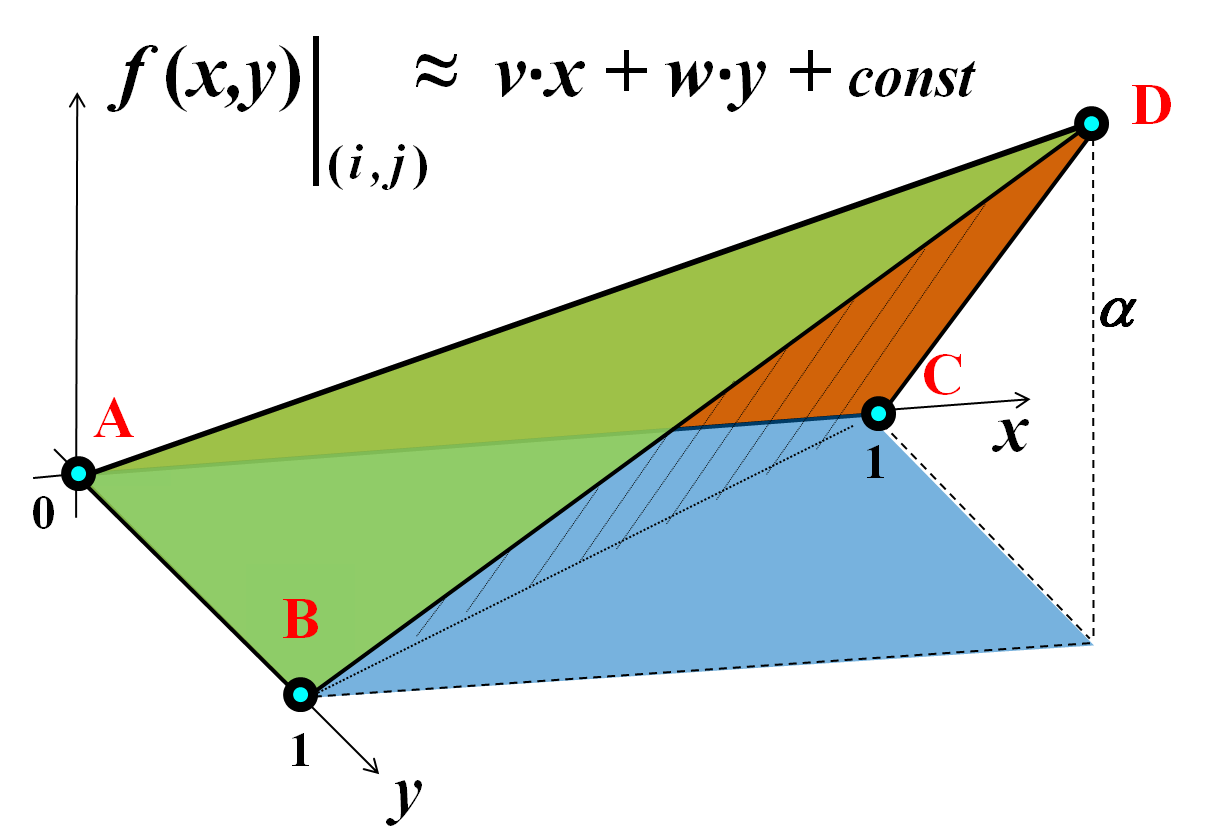} &
\includegraphics[width =0.32\textwidth]{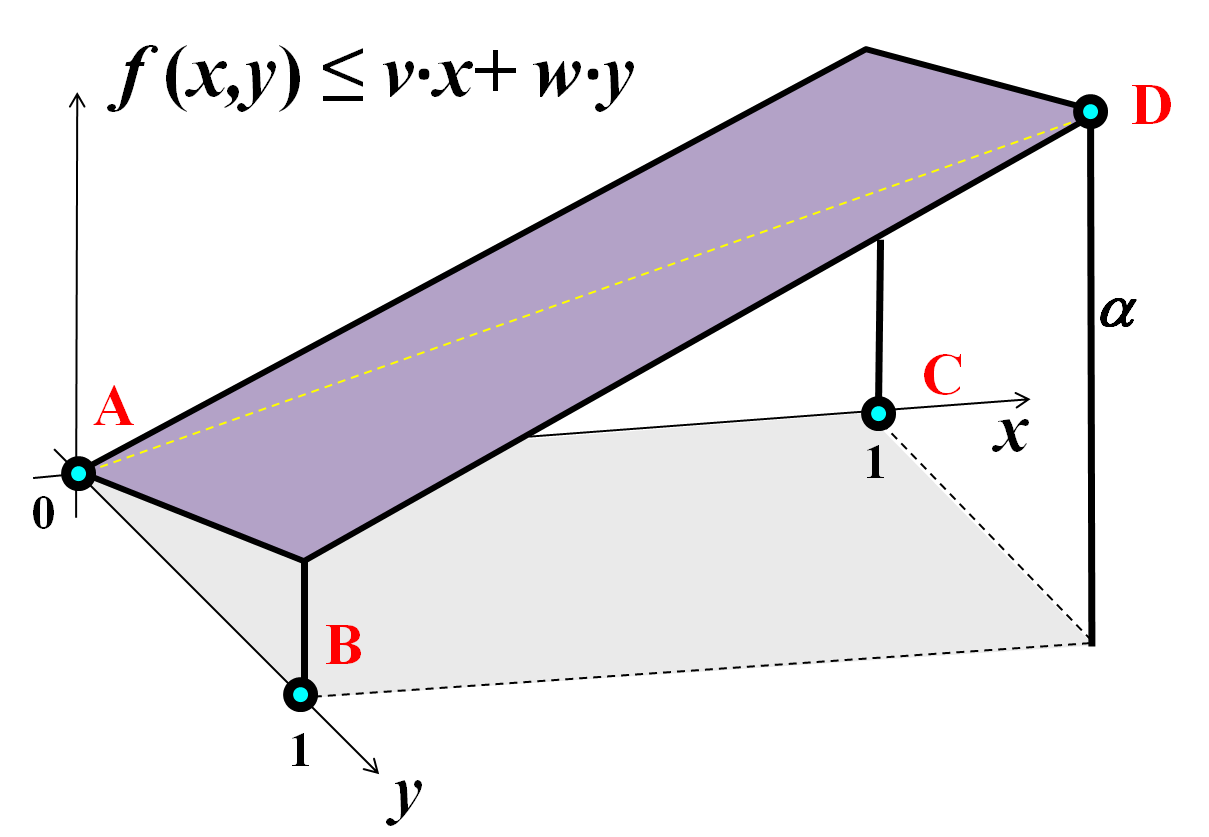} \\
(a) supermodular potential $\alpha\cdot xy$  & (b) ``Taylor'' based local linearizations & 
(c) Upper-bound linearization
\end{tabular}
\caption{Local linearization of supermodular pairwise potential $f(x,y)=\alpha\cdot xy$ for $\alpha>0$. This potential defines 
four costs $f(0,0)=f(0,1)=f(1,0)=0$ and $f(1,1)=\alpha$ at four distinct configurations of binary variables $x,y\in\{0,1\}$.
These costs can be plotted as four 3D points $A$, $B$, $C$, $D$ in (a-c). 
We need to approximate supermodular potential $f$ with a linear function $v\cdot x + w\cdot y+const$ (plane or unary potentials). 
{\bf LSA-TR:} one way to derive a local linear approximation is to take Taylor expansion of $f(x,y)=\alpha\cdot xy$ 
over relaxed variables $x,y\in[0,1]$, see the continuous plot in (a). At first, this idea may sound strange since 
there are infinitely many other continuous functions that agree with $A$, $B$, $C$, $D$ 
but have completely different derivatives, \eg $g(x,y)=\alpha\cdot x^2\sqrt{y}$. 
However, Taylor expansions of {\bf bilinear} function $f(x,y)=\alpha\cdot xy$ can be motivated  geometrically.
As shown in (b), Taylor-based local linear approximation of $f$ at any fixed integer configuration $(i,j)$ 
(\eg. blue plane at $A$, green at $B$, orange at $C$, and striped at $D$) coincides with discrete pairwise potential $f$ 
not only at point $(i,j)$ but also with two other closest integer configurations. Overall, each of those 
planes passes exactly through three out of four points $A$, $B$, $C$, $D$. {\bf LSA-AUX:}
another approach to justify a local linear approximation for non-submodular pairwise potential $f$ 
could be based on upper bounds passing through a current configuration. For example, the green or orange
planes in (b) are the tightest linear upper bounds at configurations $(0,1)$ and $(1,0)$, correspondingly. 
When current configuration is either $(0,0)$ or $(1,1)$ then one can choose either orange or green plane in (b),
or anything in-between, \eg the purple plane passing though $A$ and $D$ in (c).
\label{fig:approx}} 
\end{figure*}

\subsection{LSA-TR}\label{sec:tr}
Trust region methods are a class of iterative optimization
algorithms. In each iteration, an approximate model of the
optimization problem is constructed near the current solution $S_0$. The
model is only accurate within a small region around the current
solution called ``trust region''. The approximate model
is then globally optimized within the trust region to obtain a
candidate solution. This step is called {\em trust
  region sub-problem}.  The size of the trust region is adjusted in
each iteration based on the quality of the current
approximation. For a detailed review of trust region framework see \cite{TRreview:Yuan}.

Below we provide details for our trust region approach to the binary pairwise energy optimization 
(see pseudo-code in Algorithm \ref{alg:TR}).  The goal is to minimize $E(S)$ in \eqref{eq:en}. 
This energy can be decomposed into submodular and supermodular parts $E(S)=E^{sub}(S) + E^{sup}(S)$ such that
\begin{eqnarray*}
E^{sub}(S) & = & S^T U + S^T M^- S \\
E^{sup}(S) & = & S^T M^+ S
\end{eqnarray*}
where matrix $M^-$ with negative elements $m^-_{pq} \leq 0$ represents 
the set of submodular pairwise potentials and matrix $M^+$ with positive elements $m^+_{pq}\geq 0$ represents 
supermodular potentials. Given the current solution $S_t$ energy $E(S)$ can be approximated by submodular function
\begin{equation}\label{eq:eqTR}
E_t(S) = E^{sub}(S)  + S^T U_t + const
\end{equation}
where $U_t = 2M^+ S_t$. The last two terms in \eqref{eq:eqTR} are the first-order Taylor expansion of
supermodular part $E^{sup}(S)$. 

While the use of Taylor expansion may seem strange in 
the context of functions of integer variables, Figure \ref{fig:approx}(a,b) illustrates its geometric motivation. 
Consider individual pairwise supermodular potentials $f(x,y)$ in 
$$E^{sup}(S) = \sum_{pq} m^+_{pq}\cdot s_p s_q = \sum_{pq} f_{pq}(s_p,s_q).$$ 
Coincidentally, Taylor expansion of each relaxed supermodular potential $f(x,y)=\alpha\cdot xy$ 
produces a linear approximation (planes in b) that agrees with $f$ at three out of four 
possible discrete configurations (points A,B,C,D).

The standard trust region sub-problem is to minimize approximation $\widetilde{E}$ within the region defined by step size $d_t$ 
\begin{equation} \label{eq:constrained}
S^*= \underset{||S-S_t||<d_t}{\operatorname{argmin}} E_t(S).
\end{equation}
Hamming, $L_2$, and other useful metrics $||S-S_t||$ can be represented by a sum of unary potentials \cite{pdecut-eccv06}. 
However, optimization problem \eqref{eq:constrained} is NP-hard even for unary metrics\footnote{By a  reduction to the {\em balanced cut} problem.}\@. One can solve Lagrangian dual of \eqref{eq:constrained} by iterative sequence of graph cuts
as in \cite{kahl:13}, but the corresponding duality gap may be large and the optimum for \eqref{eq:constrained} is not guaranteed. 

Instead of \eqref{eq:constrained} we use a simpler formulation of the trust region subproblem proposed in \cite{FTR:cvpr13}.
It is based on unconstrained optimization of submodular Lagrangian 
\begin{equation} \label{eq:lagrangian}
L_t(S) = E_t(S) + \lambda_t\cdot||S-S_t||
\end{equation}
where parameter $\lambda_t$ controls the trust region size indirectly. 
Each iteration of LSA-TR solves \eqref{eq:lagrangian} for some fixed $\lambda_t$ and
adaptively changes $\lambda_t$ for the next iteration (Alg.\ref{alg:TR} line \ref{line:tau2}), 
as motivated by empirical inverse proportionality relation between
$\lambda_t$ and $d_t$ discussed in \cite{FTR:cvpr13}.

Once a candidate solution $S^*$ is obtained, the quality of the
approximation is measured using the ratio between the actual and
predicted reduction in energy. Based on this ratio, the solution is
updated in line \ref{line:tau1} and the step size (or $\lambda$) is adjusted in
line \ref{line:tau2}. It is common to set the parameter $\tau_1$ in
line \ref{line:tau1} to zero, meaning that any candidate solution that
decreases the actual energy gets accepted. The parameter $\tau_2$ in
line \ref{line:tau2} is usually set to 0.25
\cite{TRreview:Yuan}. Reduction ratio above this value corresponds to
good approximation model allowing increase in the trust region size.

\begin{algorithm}[h!]
{\small
\DontPrintSemicolon
\textbf{Initialize} \hspace{1ex} $t=0$, \hspace{1ex} $S_0$, \hspace{1ex} $\lambda_0$ \;
\textbf{Repeat} \;
\hspace*{0.2cm} \textbf{//Solve Trust Region Sub-Problem} \;
\hspace*{0.3cm} $S^* \longleftarrow \operatorname{argmin}_{S\in\{0,1\}^\Omega} L_t(S)$  \hspace{1ex} // as defined in \eqref{eq:lagrangian} 
\; \label{line:TRsolve}
\hspace*{0.3cm} $P=E_t(S_t)-E_t(S^*)$ //predicted reduction in energy\;
\hspace*{0.3cm} $R =E(S_t) - E(S^*)$ //actual reduction in energy\;
\hspace*{0.2cm} \textbf{//Update current solution} \;
\hspace*{0.3cm}$S_{t+1} \longleftarrow
\left\{
	\begin{array}{ll}
		S^* & \mbox{if } R/P>\tau_1 \\ 
		S_t & \mbox{otherwise} 
	\end{array}
\right.$\;\label{line:tau1}
\hspace*{0.2cm} \textbf{//Adjust the trust region} \;
\hspace*{0.3cm} $\lambda_{t+1} \longleftarrow
\left\{
	\begin{array}{ll}
		\lambda_t / \alpha & \mbox{if } R/P>\tau_2 \\
		\lambda_t \cdot \alpha  & \mbox{otherwise } 
	\end{array}
\right.$\; \label{line:tau2}
\textbf{Until Convergence} \;
}
\caption{{\small \textsc{General Trust Region Approach} \label{alg:TR}}
}
\end{algorithm}

\subsection{LSA-AUX}\label{sec:aux}

Bound optimization techniques are a class of iterative optimization
algorithms constructing and optimizing upper bounds, a.k.a. {\em auxiliary functions}, 
for energy $E$. It is assumed that those bounds are easier to optimize than the original energy $E$.  
Given a current solution $S_t$, the function $A_t(S) $ is an auxiliary function of $E$ if
it satisfies the following conditions:
\begin{subequations}
\begin{align}
E(S) \, &\leq \, A_t(S) \label{general-second} \\
E(S_t) &= A_t(S_t) \label{general-third} 
\end{align}
\label{Eq:Auxiliary_function_conditions}
\end{subequations}
To approximate minimization of $E$, one can iteratively minimize
a sequence of auxiliary functions:
\begin{equation}
S_{t+1} = \arg \min_{S} \; A_t(S) \, , \quad t=1, 2, \dots
\label{Eq:Iterative_bound_minimization}
\end{equation}
Using \eqref{general-second}, \eqref{general-third}, and  \eqref{Eq:Iterative_bound_minimization}, it is straightforward to prove that the 
solutions in \eqref{Eq:Iterative_bound_minimization} correspond to a sequence of decreasing energy values $E(S_t)$. Namely,
\begin{equation}
E(S_{t+1}) \leq  A_t(S_{t+1}) \leq A_t(S_t) = E(S_t).  \nonumber
\end{equation}

The main challenge in bound optimization approach is designing an appropriate auxiliary function 
satisfying conditions \eqref{general-second} and \eqref{general-third}. However, in case of integer quadratic 
optimization problem \eqref{eq:en}-\eqref{eq:iqp},  it is fairly straightforward to
design an upper bound for non-submodular energy $E(S)=E^{sub}(S) + E^{sup}(S)$.
As in Sec.\ref{sec:tr}, we do not need to approximate the submodular part $E^{sub}$
and we can easily find a linear upper bound for $E^{sup}$ as follows.

Similarly to Sec.\ref{sec:tr}, consider supermodular pairwise potentials $f(x,y)=\alpha\cdot xy$ 
for individual pairs of neighboring pixels according to 
\begin{equation}\label{eq:sum}
E^{sup}(S) = \sum_{pq} m^+_{pq}\cdot s_p s_q = \sum_{pq} f_{pq}(s_p,s_q)
\end{equation}
where each $f_{pq}$ is defined by scalar $\alpha = m^+_{pq} >0$. 
As shown in Figure \ref{fig:approx}(b,c),  each pairwise potential $f$ can be bound above by linear
function $u(x,y)$
$$ f(x,y)\leq u(x,y) := v\cdot x+w\cdot y$$
for some positive scalars $v$ and $w$.
Assuming current solution $(x,y)=(x^t,y^t)$, the table below specifies linear upper bounds (planes)
for four possible discrete configurations
\begin{center}
\begin{tabular}{|c|c|c|}
\hline
$(x^t, y^t)$ & upper bound $u(x,y)$  & plane in Fig.\ref{fig:approx}(b,c)\\
\hline
(0,0) & $\frac{\alpha}{2}x + \frac{\alpha}{2}y$   & purple \\
(0,1) & $\alpha x$                                                 &  green \\
(1,0) &  $\alpha y$                                                &  orange \\
(1,1) & $\frac{\alpha}{2}x + \frac{\alpha}{2}y$    &  purple \\
\hline
\end{tabular}
\end{center}
As clear from Fig.\ref{fig:approx}(b,c), there are many other possible linear upper bounds for pairwise terms $f$.
Interestingly, the ``permutation'' approach to high-order supermodular terms in \cite{Bilmes2005} reduces 
to linear upper bounds for $f(x,y)$ where each configuration (0,0) or (1,1) selects either orange or green 
plane randomly (depending on a permutation). Our tests showed inferior performance of such bounds for
pairwise energies. The upper bounds using purple planes for (0,0) and (1,1), as in the table, work better in practice.

\begin{figure}[t!]
\begin{center}
\includegraphics[width = 1\columnwidth]{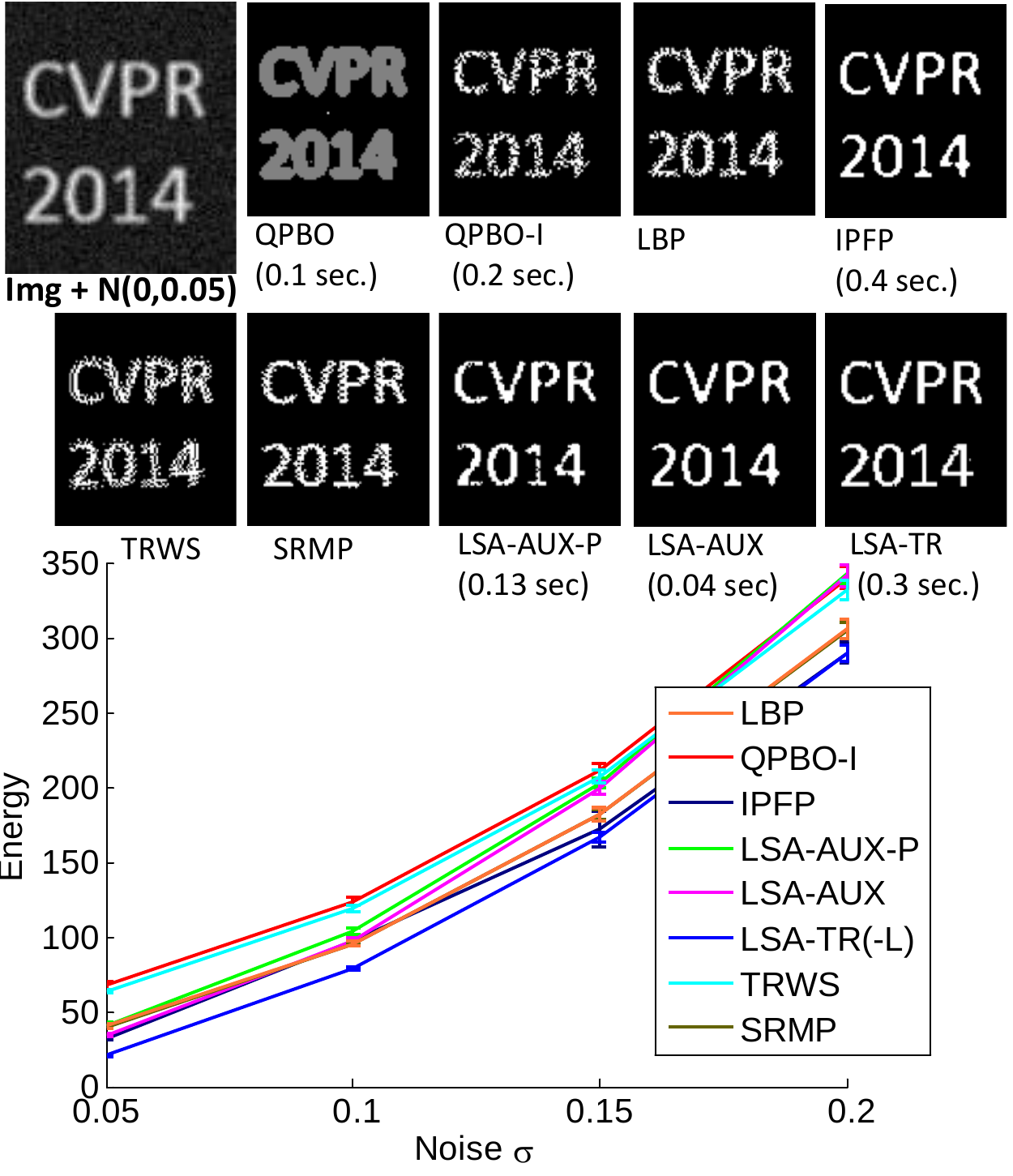}
\end{center}
\vspace{-2ex}
\caption{Binary deconvolution of an image created with a uniform $3\times3$ filter and additive Gaussian noise ($\sigma \in \{0.05,0.1,0.15,0.2\}$). No length regularization was used. We report mean energy (+/-2std.) and time as a function of noise level $\sigma$. TRWS, SRMP and LBP are run for 5000 iterations.  \label{fig:deblur}} 
\end{figure}

Summing upper bounds for all pairwise potentials $f_{pq}$ in \eqref{eq:sum} using linear terms
in this table gives an overall linear upper bound for supermodular part of energy \eqref{eq:en}
\begin{equation} \label{eq:supAux}
E^{sup}(S)\leq S^T U_t
\end{equation}
where vector $U_t = \{u^t_p|p \in \Omega \}$ consists of elements
$$
u^t_p = \sum_{q } \frac{m^+_{pq}}{2}(1+s^t_p -2 s^t_p s^t_q)
$$
and $S_t = \{s^t_p|p\in \Omega\}$ is the current solution configuration for all pixels.
Defining our auxiliary function as
\begin{equation}\label{eq:auxiliary_bound}
A_t(S) := S^T U_t + E^{sub}(S)
\end{equation}
and using inequality \eqref{eq:supAux} we satisfy condition \eqref{general-second} 
$$E(S) = E^{sup}(S)  + E^{sub}(S)  \leq  A_t(S).$$
Since $S_t^T U_t = E^{sup}(S_t)$ then our auxiliary function \eqref{eq:auxiliary_bound}
also satisfies condition \eqref{general-third}
$$E(S_t) = E^{sup}(S_t)  + E^{sub}(S_t)  =  A_t(S_t).$$
Function $A_t(S)$ is submodular. Thus, we can globally optimize it 
in each iterations guaranteeing energy decrease.

\section{Applications}\label{sec:applications}

\begin{figure}[t!]
\begin{center}
\includegraphics[width =1\columnwidth]{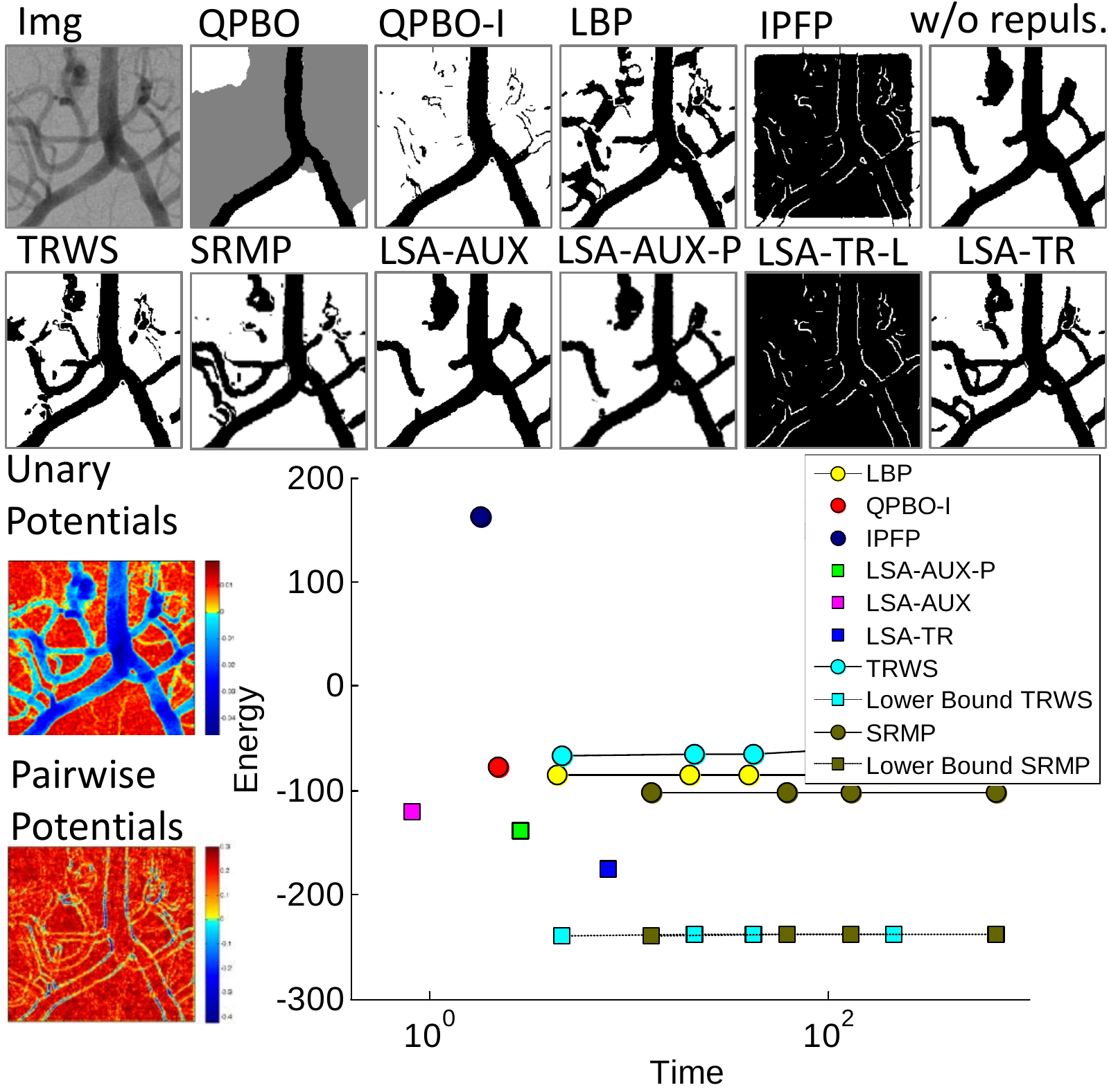}
\caption{Segmentation with repulsion and attraction. We used $\mu_{fg}$=0.4, $\mu_{bg}$=0.6, $\sigma$=0.2 for appearance, $\lambda_{reg}$=100 and c=0.06. Repulsion potentials are shown in blue and attraction - in red. \label{fig:repulsion}}
\vspace{-0.5cm}
\end{center}
\end{figure} 

Below we apply our method in several applications such as binary deconvolution, segmentation with repulsion, curvature regularization and inpainting. We report results for both LSA-TR and LSA-AUX frameworks and compare to existing state of the art methods such as QPBO \cite{rother-et-al-cvpr-2007}, LBP \cite{pearl-1982},  IPFP \cite{NIPS09LeordeanuHS09}, TRWS and SRMP \cite{GTRWS:arXiv12} in terms of energy and running time\footnote{We used {\em http://pub.ist.ac.at/$\sim$vnk/software.html} code for SRMP and {\em www.robots.ox.ac.uk/$\sim$ojw} code for QPBO, TRWS, and LBP. The corresponding version of LPB is sequential without damping.}. For the sake of completeness, and to demonstrate the advantage of non-linear submodular approximations over linear approximations, we also compare to a version of LSA-TR where  both submodular and supermodular terms are linearized, denoted by LSA-TR-L. In the following experiments, all local approximation methods, \eg IPFP, LSA-AUX, LSA-AUX-P, LSA-TR, LSA-TR-L are initialized with the entire domain assigned to the foreground. All global linearization methods, \eg TRWS, SRMP and LBP, are run for 50, 100, 1000 and 5000 iterations. For QPBO results, unlabeled pixels are shown in gray color. Running time is shown in log-scale for clarity.

\subsection{Binary Deconvolution}

\begin{figure*}[t!]
\begin{center}
\includegraphics[width = 0.8\textwidth]{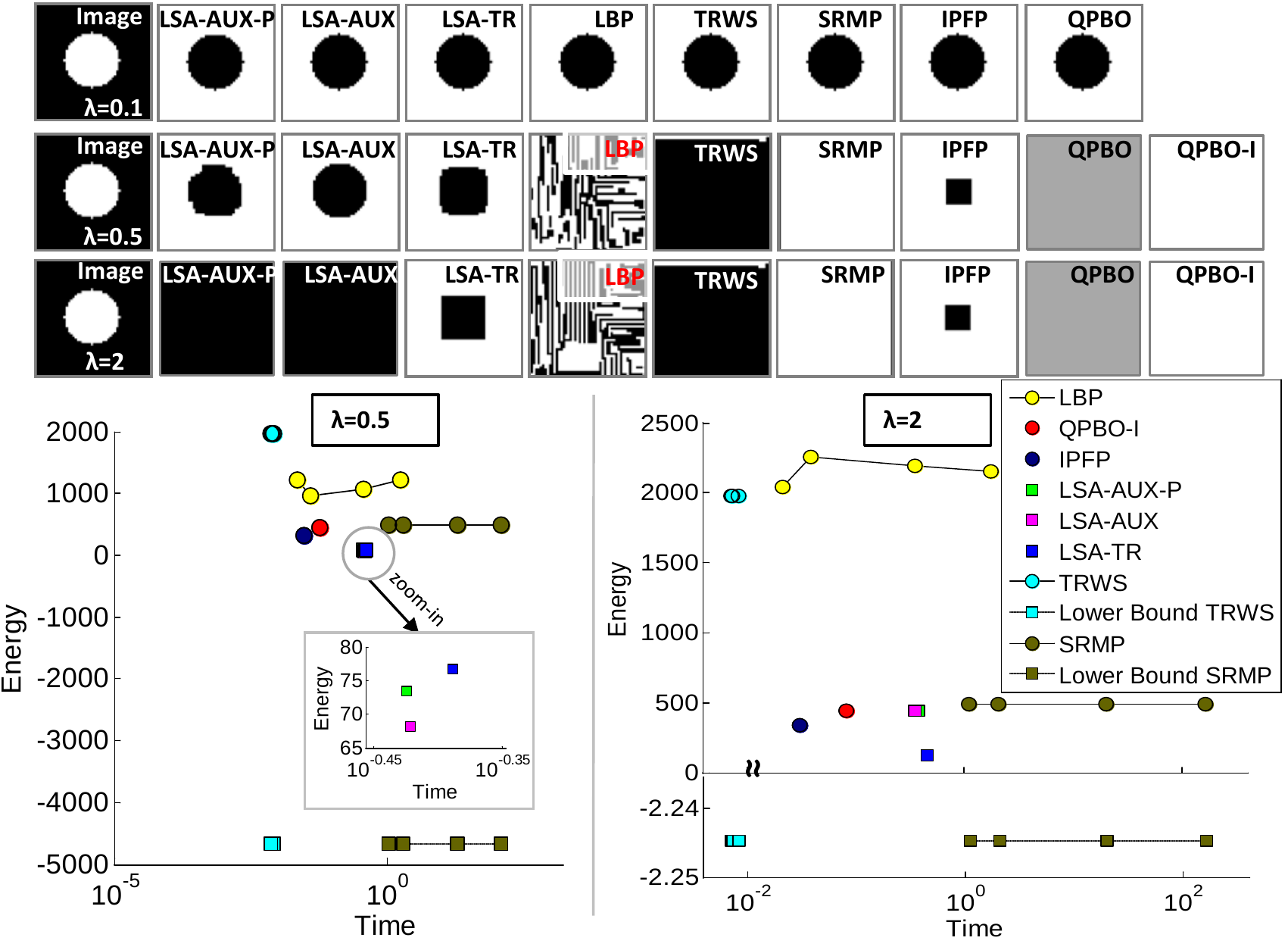}
\end{center}
\vspace{-0.3cm}
\caption{Curvature regularizer  \cite{elzehiry2010fast} is more difficult to optimize when regularizer weight is high. We show segmentation results for $\lambda_{curv}=0.1$ (top row), $\lambda_{curv}=0.5$ (middle row), $\lambda_{curv}=2$ (bottom row) as well as energy plots. We used $\mu_{fg}$ = 1, $\mu_{bg}$ = 0, $\lambda_{app}$ = 1. \label{fig:LeoCurvature_Circle}} 
\end{figure*}

Figures \ref{fig:deblur} (top-left) shows a binary image after convolution with a uniform $3\times3$ and adding Gaussian noise ($\sigma=0.05$). The goal of binary deconvolution is to recover the original binary image and the energy is defined as
\begin{equation}\label{eq:deblur}
E(S) = \sum_{p\in\Omega} (I_p - \frac{1}{9}\sum_{q\in {\cal N}_p}s_q)^2
\end{equation}

Here $ {\cal N}_p$ denotes the $3 \times 3$ neighborhood window around pixel $p$ and all pairwise interactions are supermodular. We did not use length regularization, since it would make the energy easier to optimize. Fig.~\ref{fig:deblur} demonstrates the performance of our approach (LSA-TR/LSA-AUX) and compares to  standard optimization methods such as QPBO, LBP, IPFP, TRWS and SRMP. In  this case LSA-TR-L and LSA-TR are identical since energy \eqref{eq:deblur} has no submodular pairwise terms. The bottom of Fig.~\ref{fig:deblur} 
shows the mean energy as a function of noise level $\sigma$. For each experiment the results are averaged over ten instances of random noise.  The mean time is reported for the experiments with $\sigma=0.05$.

\subsection{Segmentation with Repulsion}

In this section we consider segmentation with attraction and repulsion pairwise potentials. Adding repulsion is similar to correlation clustering \cite{Bansal02correlationclustering}, where data points either attract or repulse each other.  Using negative repulsion in segmentation can avoid the bias of submodular length regularizer to {\em short-cutting}, whereby elongated structures are shortened to avoid high length penalty. Figure~\ref{fig:repulsion} (top-left) shows an example of an angiogram image with elongated structures. We use 16-neighborhood system and the pairwise potentials are defined as follows: $$\omega(p,q) = \frac{-\Delta (p,q)+c }{\mbox{dist(p,q)}}.$$ Here $\mbox{dist(p,q)}$ denotes the distance between image pixels $p$ and $q$ and $\Delta (p,q)$ is the difference in their respective intensities (see pairwise potentials in Fig.~\ref{fig:repulsion}, bottom-left). The constant $c$ is used to make neighboring pixels with similar intensities attract and repulse otherwise. Being supermodular, repulsions potentials make the segmentation energy more difficult to optimize, but are capable to extract thin elongated structures. To demonstrate the usefulness of ``repulsion'' potentials we also show segmentation results with graph-cut a la Boykov-Jolly \cite{BJ:ICCV01} where negative pairwise potentials were removed/truncated (top-right).

\subsection{Curvature}

Below we apply our optimization method to curvature regularization. We focus on the curvature model proposed in \cite{elzehiry2010fast}. The model is defined in terms of 4-neighborhood system and accounts for 90 degrees angles. In combination with appearance terms, the model yields discrete binary energy that has both submodular and non-submodular pairwise potentials. Originally, the authors of \cite{elzehiry2010fast} proposed using QPBO for optimization of the curvature regularizer. We show that our method significantly outperforms QPBO and other state-of-the-art optimization techniques, especially with large regularizer weights.

First we deliberately choose a toy example  (white circle on a black background, see Fig.~\ref{fig:LeoCurvature_Circle}) where we know what an optimal solution should look like. When using 4-neighborhood system, as the weight of the curvature regularizer increases, the solution should minimize the number of 90 degrees angles (corners) while maximizing the appearance terms. Therefore, when the weight of curvature regularizer is high, the solution should look more like a square than a circle. Consider the segmentation results in Fig.~\ref{fig:LeoCurvature_Circle}. With low curvature weight, i.e.,~$\lambda_{curv} = 0.1$ , all compared methods perform equally well (see Fig.~\ref{fig:LeoCurvature_Circle} top row). In this case appearance data terms are strong compared to the non-submodular pairwise terms. However, when we increase the curvature weight and set $\lambda_{curv} = 0.5$ or $2$ there is a significant difference between the optimization methods both in terms of the energy and the resulting solutions (see Fig.~\ref{fig:LeoCurvature_Circle} middle and bottom). 

\begin{figure*}[t!]
\begin{center}
\includegraphics[width = 0.8\textwidth]{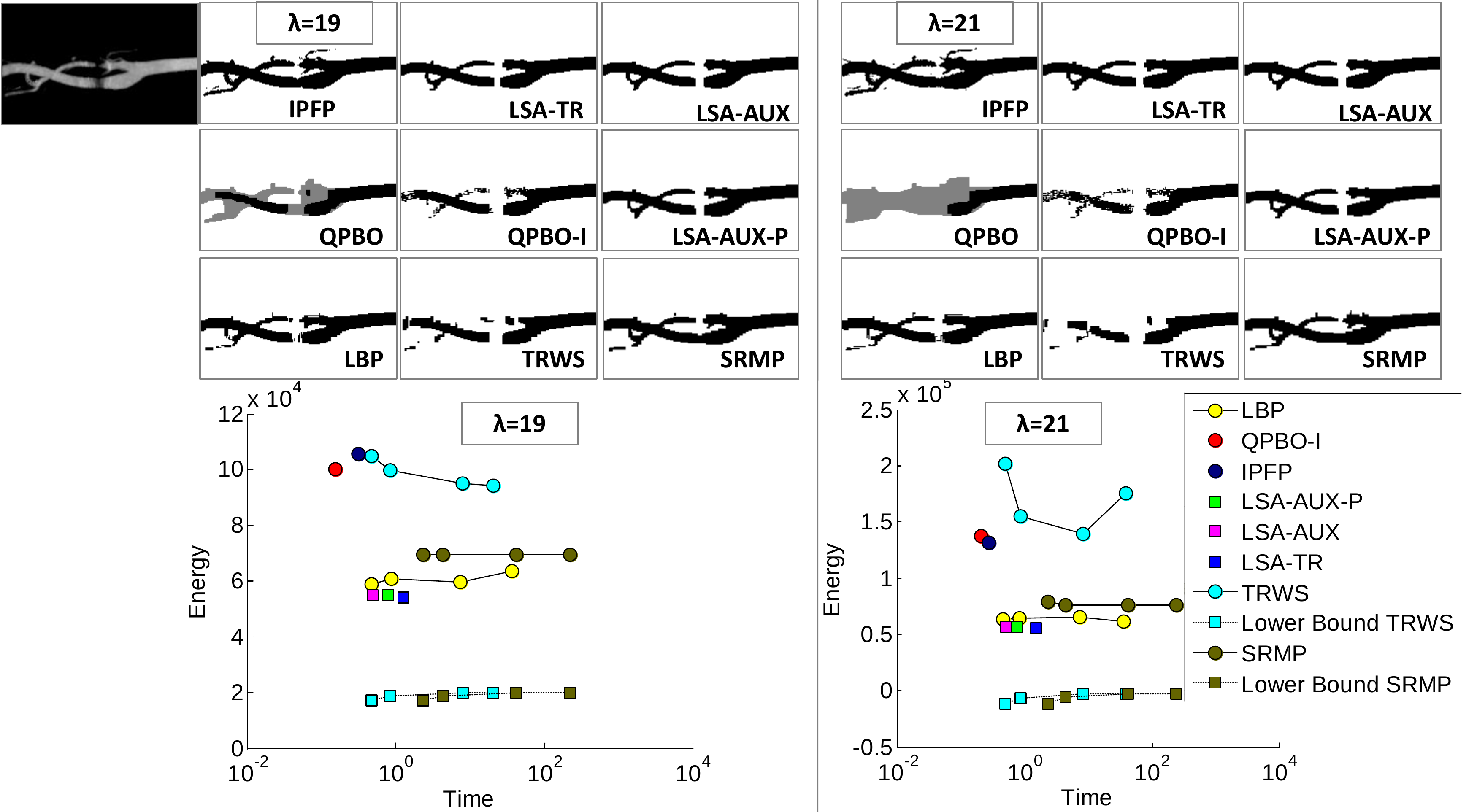}
\end{center}
\vspace{-0.5cm}
\caption{Curvature regularizer \cite{elzehiry2010fast}: we show segmentation results and energy plots for  $\lambda_{curv}$=19 (left), $\lambda_{curv}$=21 (right).  \label{fig:LeoCurvature_Stenosis}}
\vspace{-0.5cm}
\end{figure*}

Next, we selected an angiogram image example from \cite{elzehiry2010fast} and evaluate the performance\footnote{For QPBO, we only run QPBO-I and do not use other post-processing heuristics as suggested in \cite{elzehiry2010fast}, since the number of unlabeled pixel might be significant when the regularization is strong.} of the optimization methods with two values of regularizer weight $\lambda_{curv}=19$ and $\lambda_{curv}=21$ (see Fig.~\ref{fig:LeoCurvature_Stenosis}).  Although the weight $\lambda$ did not change significantly, the quality of the segmentation deteriorated for all global linearization methods, namely QPBO, TRWS, LBP. The proposed methods LSA-TR and LSA-AUX seem to be robust with respect to the weight of the supermodular part of the energy.

\subsection{Chinese Characters Inpainting}

\begin{figure}[t]
\begin{center}
\includegraphics[width = 0.6\columnwidth]{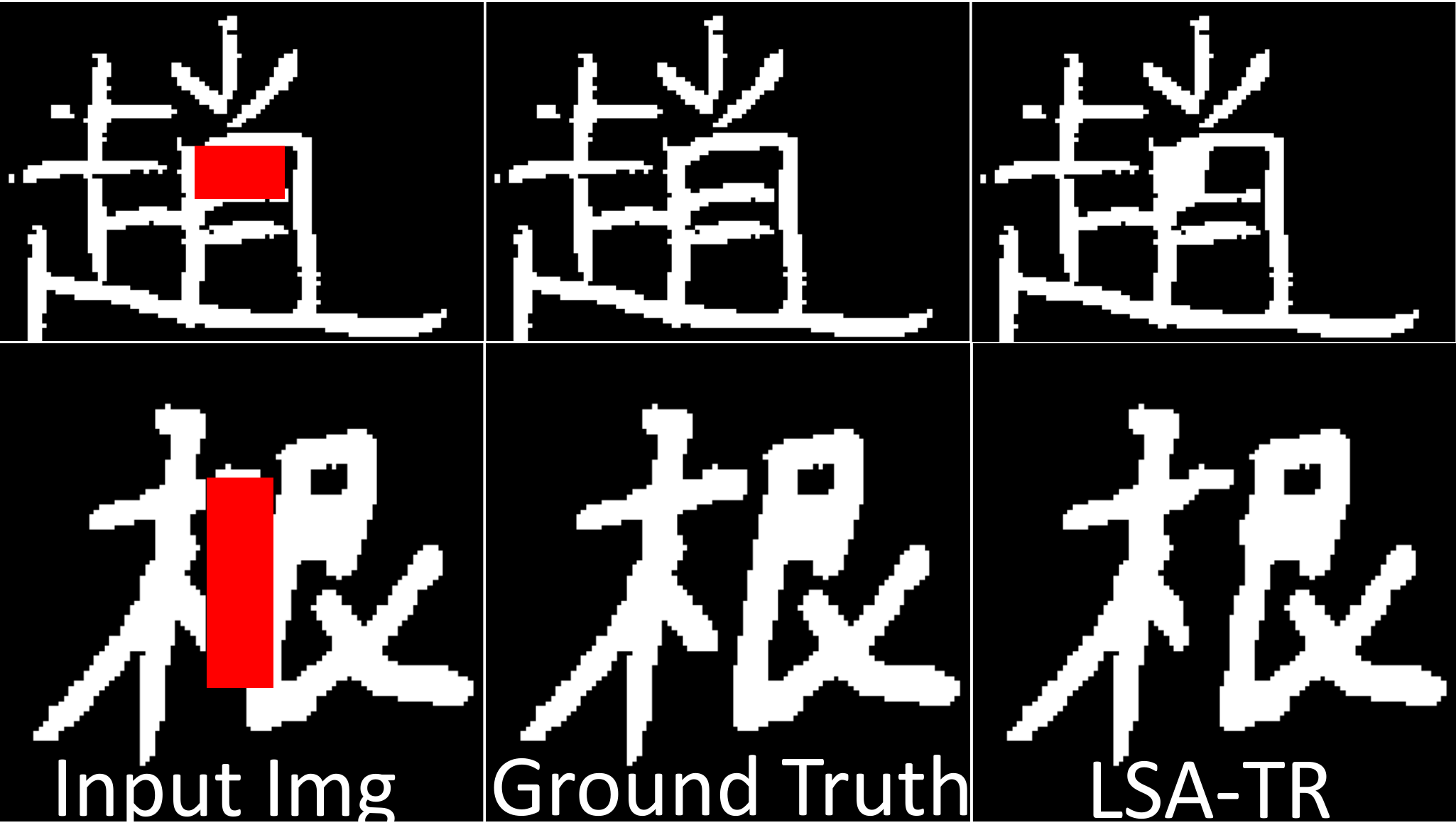}
\end{center}
\vspace{-0.1cm}
\caption{Examples of Chinese characters inpainting. \label{fig:chinese_examples}}
\vspace{-0.1cm}
\end{figure}

Below we consider the task of in-painting in binary images of Chinese characters, \emph {dtf-chinesechar} \cite{kappes-2013}. We used a set of pre-trained unary and pairwise potentials provided by the authors with the dataset. While each pixel variable has only two possible labels, the topology of the resulting graph and the non-submodularity of its pairwise potentials makes this problem challenging. Figure \ref{fig:chinese_examples} shows two examples of inpainting. Table \ref{tab:chinese} reports the performance of our LSA-TR and LSA-AUX methods on this problem and compares to other standard optimization methods reported in \cite{kappes-2013}, as well as, to {\em truncation} of non-submodular terms. 
LSA-TR is ranked second, but runs three orders of magnitudes faster.

\begin{table}[h!]
\begin{center}
\begin{small}
\begin{tabular}{|p{1.7cm}|p{1.5cm}|p{1.3cm}|c|c|}
\hline
\begin{tabular}{l} Alg.\\ Name\end{tabular} &\begin{tabular}{l} Mean\\ Runtime\end{tabular} & \begin{tabular}{l} Mean\\ Energy\end{tabular} & \begin{tabular}{l} \#best\\ /100\end{tabular} & Rank\\
\hline
MCBC&2053.89 sec &-49550.1&85&1\\
\hline
BPS (LBP)$^*$ &72.85 sec &-49537.08&18&3\\
\hline
ILP&3580.93 sec &-49536.59&8&6\\
\hline
QPBO&0.16 sec &-49501.95&0&8 \\
\hline
SA&NaN sec &-49533.08&13&4\\
\hline
TRWS&0.13 sec &-49496.84&2&7\\
\hline
TRWS-LF&2106.94 sec &-49519.44&11&5\\
\hline\hline
Truncation& 0.06 sec&-16089.2&0&8\\ 
\hline
LSA-AUX&0.30 sec&-49515.95&0&	8\\
\hline
LSA-AUX-P&0.16 sec&-49516.63&0&	8\\
\hline
LSA-TR&0.21 sec& -49547.61&35&2\\
\hline
\end{tabular}
\end{small}
\end{center}
\vspace{-0.05cm}
\caption{Chinese characters in-painting database \cite{kappes-2013}. We tested three methods (at 
the bottom) and compared with other techniques (above) reported in  \cite{kappes-2013}. 
* - To the best of our knowledge,  BPS in \cite{kappes-2013} is the basic sequential version of loopy belief-propagation 
without damping that we simply call LBP in this paper.
\label{tab:chinese}}
\vspace{-0.2cm}
\end{table}

\section{Conclusions and Future Work}\label{sec:conclusion}
There are additional applications (beyond the scope of this paper) that can benefit from efficient optimization of binary non-submodular pairwise energies. For instance, our experiments show that our approach can improve non-submodular $\alpha$-expansion and fusion moves for multilabel energies. 
Moreover, while our paper focuses on pairwise interactions, our approach naturally extends to high-order potentials that appear in computer vision problems such as curvature regularization, convexity shape prior, visibility and silhouette consistency in multi-view reconstruction.
In the companion paper \cite{curvatureFTR} we apply our method for optimization of a new highly accurate curvature regularization model. The model yields energy with triple clique interactions and our method achieves state-of-the-art performance. 

\section*{Acknowledgments}
We greatly thank V. Kolmogorov and our anonymous reviewers for their thorough feedback.
We also thank Canadian granting agency NSERC for its continued support.

{\small
\bibliographystyle{ieee}
\bibliography{arXiv_lsa2}
}

\end{document}